\documentclass[conference]{IEEEtran}
\usepackage{graphicx,times,amsmath,amssymb,multirow,subcaption,color}
\usepackage{mathrsfs}
\usepackage[justification = centering]{caption}
\usepackage[LGR,T1]{fontenc}
\usepackage[latin9]{inputenc}
\usepackage{wrapfig}
\usepackage{array,textcomp,stackrel,url,mathtools,enumerate}
\usepackage{multirow} %For multirow in tables
\usepackage{multicol} %For multicol in tables
\usepackage{booktabs,helvet,courier,float,csquotes}
\usepackage{algorithm,algcompatible}
\usepackage{flushend}
\usepackage{csquotes}
\usepackage[colorinlistoftodos]{todonotes}
\usepackage{color}
\usepackage{color,soul}
\usepackage{blindtext}
\usepackage[inline]{enumitem}
\usepackage{pdfpages}
\usepackage{pdflscape}
\usepackage{wrapfig}
\usepackage{cite}

\usepackage{tikz}  \usetikzlibrary{matrix,chains,positioning,decorations.pathreplacing,arrows}
\algnewcommand\INPUT{\item[\textbf{Input:}]}%
\algnewcommand\OUTPUT{\item[\textbf{Output:}]}%

\newcolumntype{x}[1]{>{\centering\arraybackslash\hspace{0pt}}p{#1}}
\makeatletter

\begin{document}
\title{Machine Teaching in Hierarchical Genetic Reinforcement Learning: Curriculum Design of Reward Functions for Swarm Shepherding}

%\title{A Curriculum Design of Reward Functions for Reinforcement Learning of Shepherding Behaviours}

\author{\IEEEauthorblockN{Nicholas R. Clayton}
\IEEEauthorblockA{\textit{School of Engineering \& IT} \\
\textit{University of New South Wales}\\
Canberra, Australia\\
nicholas.clayton.3001@gmail.com}
\and
\IEEEauthorblockN{Hussein Abbass}
\IEEEauthorblockA{\textit{School of Engineering \& IT} \\
\textit{University of New South Wales}\\
Canberra, Australia\\
h.abbass@adfa.edu.au}
}
\maketitle

\begin{abstract}
The design of reward functions in reinforcement learning is a human skill that comes with experience. Unfortunately, there is not any methodology in the literature that could guide a human to design the reward function or to allow a human to transfer the skills developed in designing reward functions to another human and in a systematic manner. In this paper, we use Systematic Instructional Design, an approach in human education, to engineer a machine education methodology to design reward functions for reinforcement learning. We demonstrate the methodology in designing a hierarchical genetic reinforcement learner that adopts a neural network representation to evolve a swarm controller for an agent shepherding a boids-based swarm. The results reveal that the methodology is able to guide the design of hierarchical reinforcement learners, with each model in the hierarchy learning incrementally through a multi-part reward function. The hierarchy acts as a decision fusion function that combines the individual behaviours and skills learnt by each instruction to create a smart shepherd to control the swarm. 
\end{abstract}

\begin{IEEEkeywords}
Curriculum Design, Machine Teaching, Machine Education, Reinforcement Learning, Swarm Control, Systematic Instructional Design
\end{IEEEkeywords}

\section{Introduction}\label{intro}
\IEEEPARstart{M}{achine} learning systematically applies algorithms to estimate and approximate the underlying relationships hidden in a set of features~\cite{machinelearning2015}. The role of the machine could stop at finding these relationships, as in the case of data mining, or at using them to act by actuating on the environment, as in the case of robotic control. A common form of learning for robotic control is reinforcement learning, where an agent relies on the signal it receives from its interaction with the environment or other agents to update its internal states to improve its performance over time~\cite{reinforcement1998}. The internal states of the agent classically get represented using a tabular form. In continuous domains, a function approximator such as a neural network becomes essential to compress the infinite state space an agent is embedded within. The agent needs to learn at least the parameters of the function approximator, but sometimes the learning algorithm also affords the agent with the ability to learn the structure of the function~\cite{husseinandteo2004}.

John H. Holland~\cite{adaptions1976} was the first to explore genetic reinforcement learning by designing Genetic algorithms for function optimisation, which witnessed the replacement of the classical concept of an objective function by a new concept: fitness function. A fitness function rewards a solution/chromosome based on how fit the solution is relative to the remainder of the population. Genetic reinforcement learning became popular with Whitley et al.~\cite{whitley1993genetic} paper in 1993. An agent uses selection, crossover and mutation as the three primary operators to evolve a functional approximator capable of performing the task at hand. Extending the concept to multi-objective genetic reinforcement learning, Abbass~\cite{SPANN2003} introduced the Pareto-based formulation of the neural network learning problem and applied a self-adaptive differential evolution algorithm to evolve neural architectures. The work was extended in~\cite{husseinandteo2004} to the evolution of the body and controllers of robots.

Simple reward functions demonstarted success in tasks such as game play~\cite{champandard2003ai}. Greve et al.~\cite{treward2011} explored the problem of adaptation of the continuous $T$ maze challenge with a double $T$ maze. The simulation environment takes the form of an $H$ like structure with an extra leg in the middle. Rewards are placed at the ends of the arms of the $H$, three are small values and one is a large reward. If the location of the large reward is not known, the visiting of an arm end returns a one point of reward. Once the high reward has been located, only the visiting of the high reward gives a point, until it is moved again. This result is then normalised with the number of double $T$ mazes run. This simple point based system was successful in guiding the agent. However, its objective to achieve a successful completion of the task independent of the skills being learnt by the agent. In fact, the skills to reach the arms with smaller rewards get forgotten in the simulation as the reward function switches to rewarding only when the agent reaches the arm with high reward. 

Designing reward functions for swarm control is a non-trivial task due to the complexity of the search space and behaviours needed. One example of swarm control in practice is a sheepdog shepherding a group of sheep swarming using boids rules~\cite{flocks1987}. The Shepherding problem arises when one agent or a collection of agents aim at steering swarm of agents towards a targeted location. It fits in a broad category of problems alongside crowd control, oil spill recovery, and environmental protection, due to the collect and drive nature of these tasks~\cite{lienInteractive}. The problem is modelled as a swarm control problem using Boids rules for the sheep, with the results being validated against real-world shepherding~\cite{strombom2014shepherding}. 

One factor of complexity when designing reward functions for reinforcement shepherding learners lies in the challenge of breaking down the problem into smaller sub tasks. This led to many papers on Hierarchical reinforcement learning approaches~\cite{parr1998reinforcement}. However, the design of the reward function and the hierarchy has always remained as an art that relies on human\textquoteright s experience. 

%We focus in this paper on a special class of problems known as shepherding to demonstrate the methodology. 
%
%
%While a number of attempts in the literature to model the shepherding behaviour exist, much debate remains around an agreed upon algorithm for efficient shepherding.  

%One challenge associated with these problems is the lack of data to identify the best strategies for herding. Even when data could be collected, the collection of such data is both time and resource intensive, and is mostly limited to the particular behavioral set exhibited by the species under investigation.

%
%The challenge of reinforcement learning, though, is to design appropriate reward functions to guide the behaviour of the agents. Data on a single herding animal cannot be generalised to apply to other animals, however a general methodology to design an effective reward function to find the solution can be more easily adapted to changes in animal behaviours and contexts.
%
The reward function is classically obtained by running each solution within a simulator that evaluates the performance and returns some performance metrics. This approach comes with a number of disadvantages. First, these simulations are normally very expensive causing the evolution to take enormous amount of time to complete. Second, the evolution could fail to find an appropriate solution either because the reward function does not offer sufficient gradient information to guide the evolutionary process, or the target behaviour is too complex to learn in one go. Third, even if evolution was successful in finding an appropriate solution, the resultant solution is normally a convolution of a complex behaviour that is hard to understand of explain. Fourth, the design of performance metrics to evaluate each solution are almost never discussed in the literature and are assumed to be readily exist.

The aim of this paper is to attempt overcoming these limitations. As a starting point, a need exists to design a systematic methodology to the design of the reward algorithm. We use the shepherding problem as the test case to demonstrate the methodology, whereby a (genetic) reinforcement learner can learn how to shepherd in a multi-shepherd environment. The aim of the proposed methodology is to ease out the design requirements for the fitness and reward functions in complex problems.

\section{Background Materials}
This section covers background information on shepherding and curriculum design.

\subsection{Shepherding} 

The absence of sufficient data on shepherding, the need to generate novel and optimised behaviours, and the complexity of the task motivated the work in this paper to explore the use of a reinforcement learning approach that could get adopted by a smart shepherd to collect and drive a group of sheep towards a goal. Lien and Pratt~\cite{lienInteractive} discussed the practical significance of the shepherding problem and its applications in security, civil crowd control, environmental protection, agriculture, transportation safety, education and training and in entertainment. They offered an insight into the complexity in breaking the task down into logical sub tasks due to its \enquote{highly actuated nature}.

The literature looks at defining the characteristics of both herding and shepherding animals and the way that shepherding behaviours arise from the interaction between these two. Many attempts at defining these behaviours and the mathematical models behind them have been made, each with varying degrees of success ~\cite{lien2017shepherdbehaviours,geneticsheep2016,effectivev2016,steeringRobots2017}. Successful attempts at solving the problem simplify it with the addition of obstacles and road blocks however this limits the applicability of the solution to the general case. Others have incorporated a combination of human interaction and artificial intelligence to reduce complexity~\cite{lienInteractive}. The ability for an artificial intelligence agent to solve the Shepherding problem in the general case currently does not exist.

Lien et. al.~\cite{lien2017shepherdbehaviours} defined herding as one of the four major behaviours performed by a shepherd, based on observations of sheepdogs in field tests. Here herding is performed alongside `patrolling', `collecting' and `covering'.  Str\"{o}mbom et al. in \cite{strombom2014shepherding} contributed a biologically plausible algorithm to model the behaviour of sheepdogs. The performance of the algorithm was validated against real-world data collected from field trips in the Australian environment. For herding animals, behaviour was broken into a combination of three vectors: the repulsion from the predator or \enquote{herding agent}, a natural grouping vector that mapped the tendency of herd animals to group together for safety with their nearest neighbours and a close range repulsion vector that repels herd agents away from each other to model \enquote{comfortable personal space}. Str\"{o}mbom grouped shepherding behaviour into two categories, being \enquote{collecting} herding agents into a single group and then \enquote{driving} the collected group towards the target location. 

Str\"{o}mbom\textquoteright s algorithm utilised the Boids work of C.W. Reynolds~\cite{flocks1987} in modelling the behaviour of sheep flock/herd. In particular, the model imposed three behaviours, Collision avoidance, Velocity Matching and Flock Centring. Two of these behaviours were extended to the shepherds in Str\"{o}mbom\textquoteright s model demonstrating their validity in the generation of multi shepherd dynamics. Hartman and Benes~\cite{autoBoids2006} extended the work of Reynolds a step further and introduced a dynamic leadership force that accounted for chaotic behaviour without imposing a leadership structure. 

We adopt the following notations to describe different aspects of Str\"{o}mbom\textquoteright s algorithm and the overall shepherding task.
$\Psi$ denotes the shepherd\textquoteright s position;
$\sigma$ denotes a separated sheep or sheeps;
$\Phi$ denotes the sheep global centre of mass (GCM);
$\mathscr{F}$ denotes the application of force;
$P_d$ denotes driving position;
$P_c$ denotes collecting position;
$P_G$ denotes goal location;
$\Delta$ denotes acceptable error;
$\alpha$ denotes a single shepherd from a team;
$N$ denotes the number of sheep;
$\theta$ denotes angles;
$\Sigma$ denotes reward value for an agent;
$r_a$ denotes agent to agent interaction distance;
$r_s$ denotes the shepherd agent interaction distance.

Building off of the work that Str\"{o}mbom et al. completed, Fujioka K. and Hayashi F.~\cite{effectivev2016} changed the collecting behaviour from the original Str\"{o}mbom solution to what they labelled as \textit{V-Formation Control}. In this case, when the furthest sheep distance $d_{furthest} > d_r N^{(\frac{2}{3})}$ instead of moving to the collecting point behind the furthest sheep, the shepherd acts with \textit{V control} logic in which it traces a V notch shape behind the herd, orientated towards the target location. They demonstrated that the V control performs better than the Str\"{o}mbom solution for a lower number of observable neighbours for sheep agents. In a different, but older, approach, Balch and Arkin~\cite{formations1998} looked into the control of a group in formation and then control to formation to achieve the goal. 

In a more recent piece of work Lee and Kim~\cite{steeringRobots2017}, developed an algorithm that would control individual agents as a part of a steering flock without centralised control. In their work, the steering agents only aim to steer the nearest sheep agent to the target location without consideration for other members of the flock. All agents detect the presence of the shepherd and also the target location. This is combined with a shepherds ability to collect members separated from the main flock, much like that suggested by Str\"{o}mbom. The interaction between steering agents is dominated by an algorithm that considers current actions of other steering agents and closeness of itself to the current required task (collection/patrolling or herding) but relies on no centralised control. 

Successful attempts have been made in the past to use genetic reinforcement learning to learn the shepherding behaviour. However, these attempts were limited in their utility. For example, Brule et al. \cite{geneticsheep2016} used a decision tree model alongside a genetic algorithm to develop solutions to the shepherding problem. They focused on the ability of the AI to choose a path among a predetermined set of allowable actions. The work did not attempt to learn the component skills associated with a shepherding behaviour and assumed their readiness for the genetic algorithm to synthesise. However, the attempt demonstrated the feasibility of using reinforcement learning to successfully learn this complex behaviour.

\subsection{Curriculum Design}
Machine education is a growing field of research~\cite{Leu2017ACDT,Leu2017NIPS}.
In general terms, a curriculum refers to the academic content and lessons taught in a specific course~\cite{Curriculum2015}. It typically refers to the knowledge and skills students are expected to learn. In the field of machine learning, ideas from curriculum design have been explored to structure the learning process for a machine learner. Most of the work took the form of structuring the samples needed to teach a machine learning agent. 
The aim of this paper is to use curriculum design techniques to take the target compound behaviour expected from a Reinforcement learning agent and break it down into meaningful and essential behaviours and skills, then systematically design a reward function(s) targeted at learning each skill and sub-skill.

The Dick and Carey Model for Systematic Instructional Design (SID) is an example of a curriculum design model that has been around since 1978 \cite{dijkstra2013instructional}. It was intended to be used by people developing things like textbooks, workshops and computer-based training with a specific aim of converting an aim, in this paper shepherding, into a set of instructions ready for implementation. This model consists of ten steps~\cite{dick2009systematic} listed below and while these broadly apply at developing instructions for human learners, we adapt the model to develop instructions (rewards) for a reinforcement learner. 

The ten steps of the Dick and Carey Model are:
\begin{itemize}
\item Identify an Instructional Goal
\item Conduct and Instructional Analysis
\item Identify Entry Behaviours and Characteristics
\item Write Performance Objectives
\item Develop Criterion-Referenced Assessments
\item Develop Instructional Strategy
\item Develop and/or Select Instruction
\item Design and Conduct the Formative evaluation
\item Revise the Instruction
\item Conduct the Summative Evaluation
\end{itemize}

Other models for instructional design could be used such as \textit{Analyse, Design, Develop, Implement and Evaluate} (ADDIE) model. ADDIE is a conceptual way for breaking down requirements to a series of repeatable steps that can be broadened or focused as required~\cite{branch2009instructional}. The mantra of the model and its specific application largely differs from one context to another. However, Dick and Carey\textquoteright s SID model seems to encapsulate the ADDIE model as well and as such, we use the SID model in this paper. 

\section{Methodology}

We adopt the Self-Adaptive Pareto Artificial Neural Network algorithm (SPANN)~\cite{PDE2001,PDE22001}, which evolves a neural network using the Pareto-frontier Differential Evolution (PDE) algorithm~\cite{PDE2001}. In particular, we use the SPANN-R variation of the algorithm~\cite{husseinandteo2004}, which is more suitable for reinforcement learning tasks whereby back propagation is removed and a repair function is used to evolve the crossover and mutation rates. The general steps for the SPANN-R Algorithm are shown in Algorithm~\ref{algh1}.

\begin{algorithm}[h]
    \caption{SPANN-R~\cite{husseinandteo2004}}\label{algh1}
  \begin{algorithmic}[1]   
     \INPUT: $\Omega_{G - 1}$.
    \OUTPUT: $\Omega_{G}$.
    \REPEAT
        \STATE Select all non-dominated set  
        \IF {The number of networks in the non-dominated set is less than three networks}
            \STATE Add second layer of non-dominated set by excluding the previously non-dominated solutions
            \STATE Keep doing the previous step until there are more than three solutions selected 
           \ENDIF
        \STATE Delete all remaining networks
        \REPEAT 
            \STATE Select a network to be a parent $\alpha_1$
            \STATE Select two supporting parents $ \alpha_2 \text{ and } \alpha_3$
            \STATE Select a random network connection $z$
            \STATE \textbf{Crossover} - conduct element wise with a uniform chance and on element $z$
            \STATE \textbf{Mutation} - conduct with a Uniform Chance
        \UNTIL Population has been filled with new networks
     \UNTIL All generations have been completed
  \end{algorithmic}
\end{algorithm}

The ten steps of the Dick and Carey systematic instructional design model are used in this section to guide the methodology for designing the reward functions and the hierarchy for a hierarchical reinforcement learner.

\begin{enumerate}

\item \textbf{Identify an Instructional Goal: }
In the case of a human-based learning curriculum, this step involves the decision of the broader topic to be taught. For this paper, Shepherding is the instructional goal. We use the Str\"{o}mbom algorithm as the basic model to guide the design of overall expected behaviour. While the machine learner is focusing on controlling the shepherd, the control of the sheep will be pre-scripted similar to the original Str\"{o}mbom model and was adopted from a code developed in \cite{uavShepherd2017}.

\item \textbf{Conduct Instructional Analysis: }
This is probably the greatest contribution that the adaptation of curriculum design have to the design of the reward algorithm. This step looks at breaking the instructional goal down into the step-by-step behaviours that a student needs to exhibit to reach the goal. For our task, this step breaks the Shepherding goal down into two core behaviours representing the two basic skills required for shepherding: collection and driving. These skills need to be broken down into further sub-skills. The criteria to stop the decomposition is when a sub-skill is sufficiently defined such that it is completely measurable mathematically to confirm if the skill has been learnt. The overall decomposition to achieve shepherding is given in the hierarchy charts in Figure~\ref{ann:heirarchy}. Each sub-skill is then transformed into a concrete mathematical expression representing the success metric. The algorithms including these mathematical expressions are shown in Algorithms~\ref{alg:collect} and~\ref{alg:drive}. These mathematical expressions would differ from one problem to another. However, the methodology guides the designer through the decomposition to continue decomposing the problem until concrete mathematical metrics could be designed.

\begin{figure}[h!]
\includegraphics[width=0.8\linewidth]{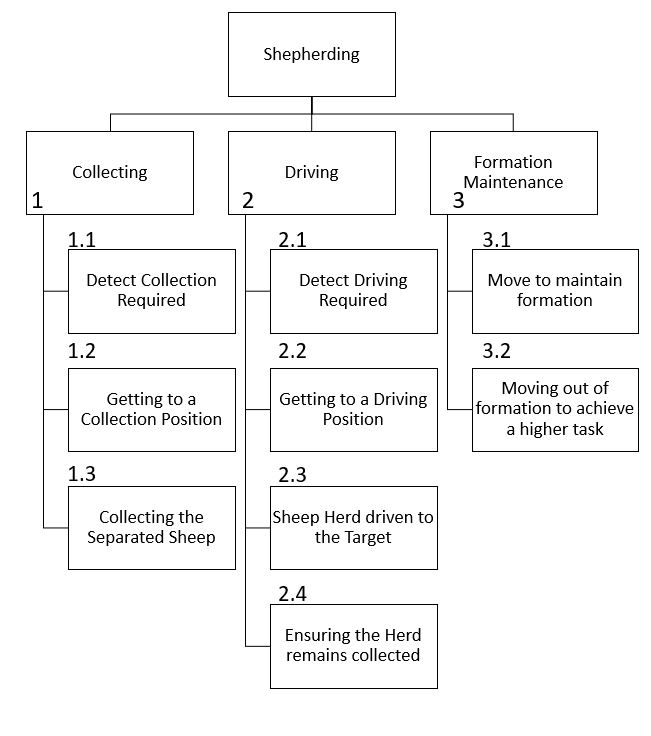}
\caption{Hierarchy Chart}\label{ann:heirarchy}
\end{figure}

\begin{algorithm}[h]
    \caption{Instruction Set for the Collection Reward}\label{alg:collect}
  \begin{algorithmic}[1]   
     \INPUT: ${State_{t - 1}, State_{t}}$.
    \OUTPUT: $\Sigma_t$.
    \STATE  Reset reward.
    \IF {Furthest Sheep is outside of herd distance}
        \STATE Award $C(t)$
    \ENDIF
    \STATE Determine $\hat{\Psi}$
    \STATE Determine $\hat{P_c}$
    \IF {($\hat{\Psi}   \pm  \Delta \theta) = \hat{P_c}$}
        \STATE Award ($\Delta \theta - ||\hat{\Psi} - \hat{P_c}||$
    \ELSE
        \STATE Punish $||\hat{\Psi} - \hat{P_c}||$
    \ENDIF
    \IF {$||\Psi - P_c||_{t} < ||\Psi - P_c||_{t - 1}$} 
        \STATE Reward $||\Psi - P_c||_{t} - ||\Psi - P_c||_{t - 1}$
    \ELSE
        \STATE Punish $2 \times (||\Psi - P_c||_{t-1} - ||\Psi - P_c||_{t})$
    \ENDIF
    \IF $|| \Psi - P_c|| > \Delta \Psi$ and $\vec{\mathscr{F}}$ on sheep
        \STATE Punish $U(t)$
    \ENDIF
    \IF {$\Psi \pm \Delta = P_c$}
        \STATE Reward $(\Delta - ||P_c - \Psi)||$
    \ENDIF
    \STATE Calculate $\hat{\sigma}$
    \STATE Calculate $\hat{\Phi}$
    \IF {($\hat{\sigma} \pm  \Delta \theta) = \hat{\Phi}$}
        \STATE Award ($\Delta \theta - ||\hat{\sigma} - \hat{\Phi}||$
    \ELSE
        \STATE Punish $||\hat{\sigma} - \hat{\Phi}||$
    \ENDIF
    \IF {$||\sigma - \Phi||_{t} < ||\sigma - \Phi||_{t - 1}$} 
        \STATE Reward $2\times (||\sigma - \Phi||_{t} - ||\sigma - \Phi||_{t - 1})$
    \ELSE
        \STATE Punish $4 \times (||\sigma - \Phi||_{t-1} - ||\sigma - \Phi||_{t})$
    \ENDIF
    \IF {$\mathscr{F}_{\Psi \rightarrow \sigma}$}
        \STATE Award $CF(t)$
    \ENDIF
  \end{algorithmic}
\end{algorithm}

\begin{algorithm}[h]
    \caption{Instruction Set for the Driving Reward}\label{alg:drive}
  \begin{algorithmic}[1]   
     \INPUT: ${State_{t - 1}, State_{t}}$.
    \OUTPUT: $\Sigma_t$.
    \STATE  Reset reward.
    \IF {Furthest Sheep is inside of herd distance}
        \STATE Award $D(t)$
    \ENDIF
    \STATE Determine $\hat{\Psi}$
    \STATE Determine $\hat{P_d}$
    \IF {($\hat{\Psi}   \pm  \Delta \theta) = \hat{P_d} $}
        \STATE Award ($\Delta \theta - ||\hat{\Psi} - \hat{P_d}||$
    \ELSE
        \STATE Punish $||\hat{\Psi} - \hat{P_d}||$
    \ENDIF
    \IF {$||\Psi - P_d||_{t} < ||\Psi - P_d||_{t - 1}$} 
        \STATE Reward $||\Psi - P_d||_{t} - ||\Psi - P_d||_{t - 1}$
    \ELSE
        \STATE Punish $2 \times (||\Psi - P_d||_{t-1} - ||\Psi - P_d||_{t})$
    \ENDIF
    \IF $|| \Psi - P_d|| > \Delta \Psi$ and $\vec{\mathscr{F}}$ on sheep
        \STATE Punish $U(t)$
    \ENDIF
    \IF {$\Psi \pm \Delta = P_d$}
        \STATE Reward $(\Delta - ||P_d - \Psi)||$
    \ENDIF
    \STATE Calculate $\hat{\Phi}$
    \STATE Calculate $\hat{P_G}$
    \IF {($\hat{\Phi} \pm  \Delta \theta) = \hat{P_G}$}
        \STATE Award ($\Delta \theta - ||\hat{\Phi} - \hat{P_G}||$
    \ELSE
        \STATE Punish $||\hat{\Phi} - \hat{P_G}||$
    \ENDIF
    \IF {$||\Phi - P_G||_{t} < ||\Phi - P_G||_{t - 1}$} 
        \STATE Reward $2\times (||\Phi - P_G||_{t} - ||\Phi - P_G||_{t - 1})$
    \ELSE
        \STATE Punish $4 \times (||\Phi - P_G||_{t-1} - ||\Phi - P_G||_{t})$
    \ENDIF
    \IF {$\mathscr{F}_{\Psi \rightarrow \Phi}$}
        \STATE Award $DF(t)$
    \ENDIF
  \end{algorithmic}
\end{algorithm}

\item \textbf{Identify Entry Behaviours and Characteristics: }
The identification of entry Behaviours in the original Dick and Carey Model looks at understanding the context in which students come to the instruction, and identifying the prerequisites, knowledge skills and attitudes required by students before undertaking the newly developed instruction. In adapting the concept to Reinforcement learning, we need to define the state for agents, identifying the required inputs and realistic outputs.

Agents in this paper are single layer Artificial Neural Networks with a maximum hidden layer of 20 nodes. The initial state of the network is defined at the start of the simulation where the original population of agents are given network weighting between -1 and 1 according to a normal distribution. A population consists of 50 agents with a simulation running a series of 250 generations. 

For agents learning to shepherd, the inputs are the relative positions of: the shepherd to the GCM, the shepherd to the furthest sheep, shepherd to the target, the GCM to the target and a bias unit with a constant input of 1.

For each agent that, the network was controlling two outputs: one defining a radian direction between $ 0 $ and $2\pi$, and a second representing a scalar speed value that would vary the size of the step that the shepherd would take. 

\item \textbf{Write Performance Objectives: }
The development of performance objectives in the Dick and Carey Model is a set of statements that define what a student will be able to do when the course has been completed (ie graduate attributes). The performance objectives need to be stated unambiguously in plain English such that they could be transformed into mathematical expressions for assessment in the next step. For shepherding, the following performance objectives stem from the task definition and the instructional analysis step:
\begin{enumerate}
\item Collecting
\begin{enumerate}
\item The Shepherd is able to detect from the simulation environment that collecting of agents is required. 
\item The Shepherd is able to move to a collecting position without causing unnecessary disruption of otherwise separated or already clustered sheep
\item The Shepherd is able to identify and guide all separated sheep back to a central herd position
\end{enumerate}
\item Driving
\begin{enumerate}
\item Detect that the conditions for Driving have been met
\item Move to a driving position without causing herd spread or by pushing the herd away
\item Guide the Herd Center of Mass to the target location.
\end{enumerate}
\end{enumerate}

\item \textbf{Develop Criterion Based Assessments: }
Criterion based assessments look to take the performance objectives and translate them into measurable assessments. Due to the way that a reinforcement learner learns from repeating the task a significant number of times, the adapting of this step saw the generation of the simulation environments. Figure~\ref{fig:environs} shows the two simulation environments that were used for the three basic behaviours. These simulation environments were later adapted further to conduct multi-behaviour testing at the Summative assessment stage which will be discussed later. 

\begin{figure*}[t]
\centering
\includegraphics[width = 0.95\textwidth]{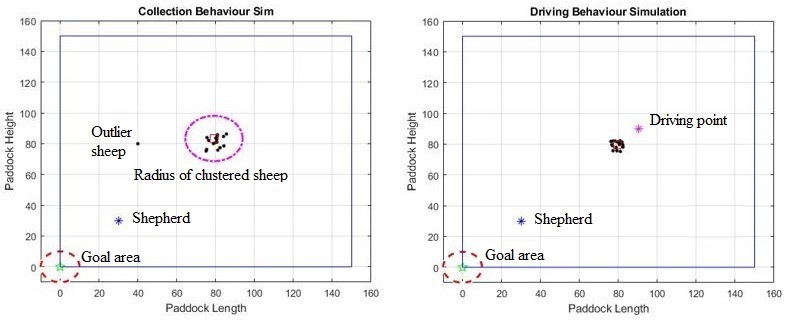}
\caption{The two simulation environments that were used to train  for collection (on left) and for driving (on right).}
\label{fig:environs}
\end{figure*}

\item \textbf{Develop Instructional Strategy: }
In order to define instructional strategy, the Dick and Carey Model looks to identify the method of instructional delivery. This includes things such as presentation of information, practice of skills, feedback to students, testing and follow up activities. In the context of reinforcement learning, this step translates into the simulation strategy. We used 250 generations, a population size of 50, and 8 selected parents. The output of simulation was a fitness graph showing the maximum, average and minimum fitness of the agents across each generation as well as recordings of the performance of key individuals at preset generations throughout the simulation in the form of a video recording. These outputs were compared across 10 simulations for each behaviour and the results were graphed, while statistics were tabulated and reported in the results section.

\item \textbf{Develop and/or Select Instruction: }
Development or selection of instruction in the classical case refers to executing the strategy so as to produce learning packages, tests, learning materials and instructor guides. For the reinforcement learning model, this step represents the development of the Reward algorithm for each of the behaviour classes; that is, the transformation of the mathematical expressions in Algorithms~\ref{alg:collect} and~\ref{alg:drive} into executable code.

\item \textbf{Design and Conduct Formative Assessment: }
The conduct of formative assessment saw the reinforcement learner attempt to learn the skills in isolated environments with the reward algorithms that had been developed for the corresponding skill. This was then followed the simulation strategy generated as a part of the Instructional strategy stage and the outputs of this stage are displayed graphically alongside the statistics data in the results section.

\item \textbf{Revise Instruction: }
The revision of instruction, or in this case the revision of the reward design, occurred concurrently throughout the conduct of formative assessment. In this way, as the running of simulations allowed for the identification of patterns that were not desirable, it meant that small changes in the ways that reward was implemented could be made.

\item \textbf{Conduct of Summative Assessments: }
The conduct of summative assessment was executed in two stages and looked to provide insight in how well the skills could be learned concurrently. The first assessment looked at the ability for the AI to learn to drive and collect at the same time. The output of this allowed an informed statement about the nature of the skills and whether they were constructive or destructive skills sets. This was then compared with the baseline model to show whether or not an improvement when using the systematic methodology was experienced.

\end{enumerate}

\textbf{Final Reward Algorithms: }
At the completion of the ten steps, a reward algorithm for each of the two core behaviours had been developed as well as a meaningful way of linking these behaviours in order to learn skills simultaneously.

\section{Results and Discussion}

\subsection{Baseline Performance}

We adopt an intuitive reward function to baseline the performance of the proposed methodology. The aim of the baseline reward, $\Sigma $, is to reward and punish the performance on the agent based on five performance metrics:
\begin{enumerate}
\item Movement towards the driving location;
\item Movement towards the GCM;
\item Changes in the average spread;
\item Changes to the distance the GCM is from the target location;
\item A final state reward component that combines the final state of the shepherds distance to the Sheep GCM ($||\Psi - \Phi||$), the sheep GCM distance to the target ($||\Psi - P_G||$) and the distance of the furthest sheep from the GCM ($||\sigma||$). 
\end{enumerate}

These are combined together with a reward starting value ($\tau$) to generate an overall fitness score ($\mu$ ) for a single agent\textquoteright s performance in the simulation as shown in Equation \ref{eqn:fitness}.

\begin{equation}
\label{eqn:fitness}
\Sigma = \tau + \Sigma - 4 \times (||\Psi - \Phi||+ ||\Psi - P_G|| + ||\sigma|| )
\end{equation}

This baseline reward function achieved a 20\% success rate. Major shortfalls in this reward function is the lack of incentive for a shepherd to collect agents separated from the main flock and the fact that the reward or punishment for the performance is given in discrete steps and doesn\textquoteright t scale with the size of the change being made.

\subsection{Formative and Summative Assessments}

\begin{figure}[H]
\centering
\includegraphics[width = 0.75\linewidth]{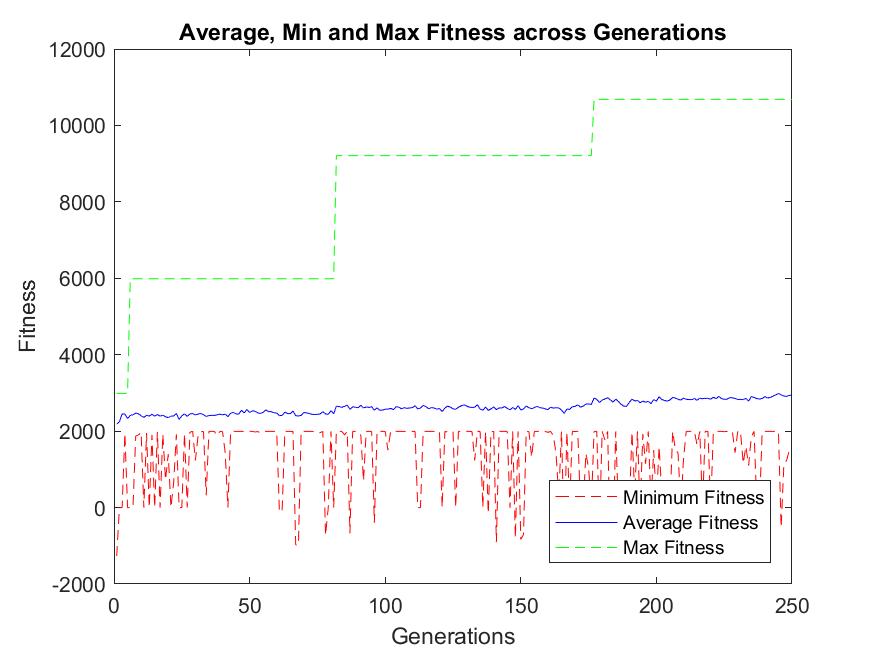}
\includegraphics[width = 0.65\linewidth]{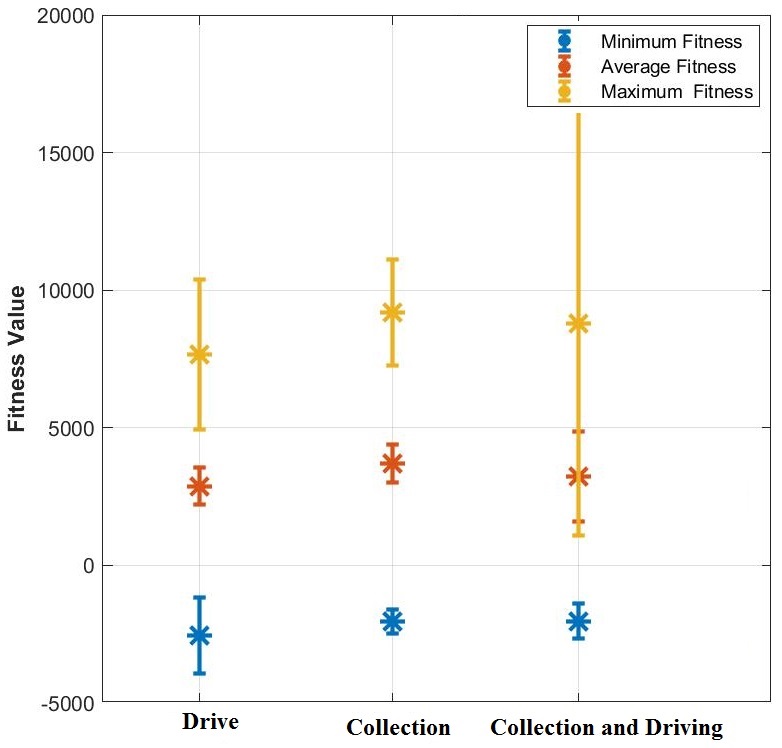}
\centering
\caption{A Typical fitness evolution throughout a simulation for a driving behaviour (top) and a statistical summary of the evolutionary results across all behaviours (bottom).}
\label{fig:typical}
\end{figure}

This section presents the quantitative and trend information for formative and summative assessments. Figure~\ref{fig:typical} shows a typical fitness model showing the evolution of agents across the 250 generations looking specifically at the minimum, maximum and average agent. This trend was consistent across all simulations.

Table \ref{tab:stats} shows the statistical variations and  success rate for each of the behaviour simulations.

\begin{table}[H]
\centering
\begin{tabular}{|c|c|c|c|c|}
\hline
\textbf{Skill}    & \textbf{Min} & \textbf{Avg $\pm$ Std Dev} & \textbf{Max} & \textbf{Success} \\ 
 &  &  &  & \textbf{Rate} \\ \hline
Collecting        & -2045.8           & 3710.2 $\pm$ 696.69               & 9188.1            & 90\%                  \\ \hline
Driving           & -2513.78          & 2879.6 $\pm$ 679.61               & 7661.2            & 80\%                  \\ \hline
Collect $+$ Drive & -2027.2           & 3236.8 $\pm$ 1630.36              & 8801.5            & 50\%                  \\ \hline
\end{tabular}
\caption{The mean values of min, average and max fitness values across different simulations.}
\label{tab:stats}
\end{table}

When compared to baseline, the results for formative assessment exhibit a higher success rate for the individual skills. Nevertheless, it is important to recognise that a skill such as driving the herd will only work when the herd is clustered. As such, the formative assessment for the individual skills are not suited for the variety of contexts that the task may exhibit. To have a better comparison against the baseline, summative assessment of collecting and driving is conducted, whereby the success rate is more than double the corresponding success rate for the baseline. While the success rate of learning the skills individually was at a higher success rate than learning both skills concurrently, both were still learnt at a significantly greater rate.

\begin{figure}[t] 
\centering
 \includegraphics[width=0.85\linewidth]{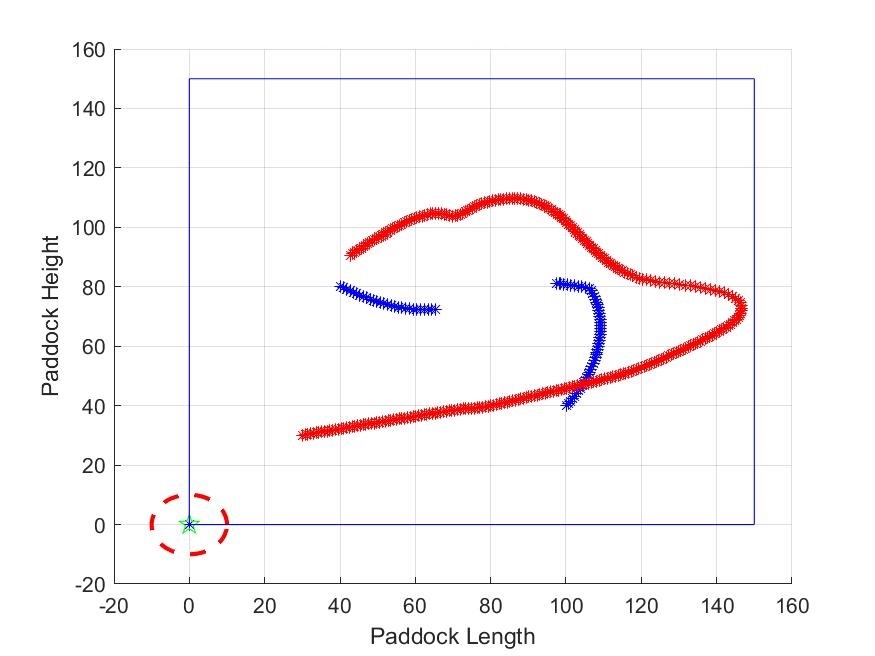}
 \includegraphics[width=0.85\linewidth]{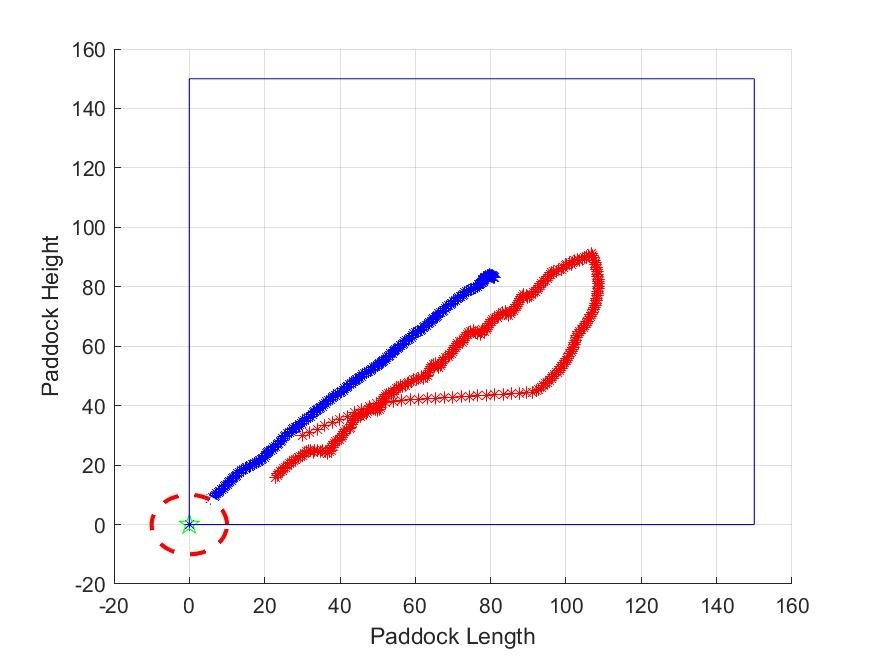}
\centering
 \caption{Foot print of the Shepherd (red) and centre of mass of the herd or the furthest sheep (blue) for collection (top) and driving (bottom). }\label{fig:driveheat}
\end{figure}

Considering that the sheep behaviour is modelled on Str\"{o}mbom\textquoteright s algorithm there is some insight to be taken in comparing the learned solution that was developed by the AI and the hard coded solution that was output by Str\"{o}mbom\textquoteright s work. Figure~\ref{fig:driveheat} show heat maps of the individual behaviours and show the distinctive differences between the driving and collecting behaviours. Similar observations could be made in Str\"{o}mbom\textquoteright s own algorithm, in that he was able to distinctly differentiate between the conduct of collection and driving. Visual inspection of the simulation running for the simultaneous learning of the collecting and driving skills show obvious sharp turns indicating a distinct change in behaviour. In this way it was insightful to see similar solutions developed and provides further validation to Str\"{o}mbom\textquoteright s interpretation of the problem.

Learning the individual behaviors independently demonstrated high success rate. When learning both skills concurrently, the success rate dropped, while was still more than double the corresponding success rate in the baseline reward function. From human learning perspective, this could have been attributed to human\textquoteright s limited cognition and cognitive load. For a neural network, this could be attributed to two main primary contributing factors. One is the larger space as discussed by Lien et al.~\cite{lienInteractive}. 

The second is the interaction effect of both skills when learnt concurrently. How to reward the agent when it is performing better on one skill than another and will an equal weighting work for this problem or different weighting is important? These are fundamental questions that have impacted the variance obtained in the success rate of different evolutionary robotics for centuries. 

The machine teaching solution opts for learning individual skills, which comes with two primary advantages. The first is the smaller search space that the machine learner needs to work on, which increases the success rate as we demonstrated above. The second is having a model specialised on each skill promotes modularity, transparency and explainability of the overall aggregate behaviour of the agent. Moreover, the scoped behavioral set associated with each individual skills allows for a better a validation of performance and diagnosis of when the learner succeeds or fails to learn.

\section{Conclusion and Future Work}

We proposed a machine teaching methodology to assist in structuring the process for designing the reward function for genetic reinforcement learners. The shepherding problem was used as the use case to demonstrate the methodology and showcase the challenges associated with the influence of the reward function design and possible interaction that the different skills an agent learns could have on the performance of reinforcement learning.

Our future work will expand the behavioural set with more complex behaviours to assist the machine teaching methodology to create machine learners with richer skills and behaviors. The use of machine teaching as a design tool to offer transparent and explainable modular neural networks is another direction that we are currently exploring. Moreover, the shepherding problem is a good testbed for our objectives because of the complex dynamics that exist in the interaction between the sheepdog and the sheep. Shepherding is only one of the four herding behaviours that was identified in \cite{lien2017shepherdbehaviours}. We will expand the work to cover the other three herding behaviours using a reinforcement learning approach.

\bibliographystyle{myIEEEtran}

\end{document}